# Autonomous Laparoscope Control through Unified Mechanics-Based Representation of Multimodal Intraoperative Information

Xiaojian Li, Jin Fang, Yudong Shi, Xilin Xiao, Kai Yan, Kang Min, Ling Li, Hua Tang, and Hangjie Mo

***Abstract*—Laparoscope-holding robots can provide surgeons with a stable laparoscopic field of view (FOV) and reduce the burden on human assistants. To maintain an ideal intraoperative FOV, the robot must continuously adjust the laparoscope pose according to intraoperative information. However, intraoperative multimodal signals, such as position, force/torque, and images, differ markedly in physical meaning and units, making it difficult to build a unified representation and to generate control commands that can be used directly for laparoscope control. To address this issue, we propose a laparoscope-holding robot control method based on unified mechanics modeling of multimodal information. First, we design mapping strategies for multiple intraoperative sources, including position, force/torque, and images, and unify them into an equivalent-wrench representation in the operational space. Then, using a task-priority scheme, we inject the wrenches into the task space and the null space, respectively, and synthesize laparoscope control commands via task-priority projection, thereby achieving consistent representation and coordinated fusion of multimodal information within a single framework. Finally, taking the intraoperative remote center of motion (RCM) position, force/torque sensor readings, and laparoscopic images as examples, we construct an RCM-constraint wrench to enforce the RCM geometric constraint and reduce the contact force at the trocar site, a laparoscope-manipulation wrench to enable compliant dragging, and an instrument-tracking wrench to achieve autonomous visual tracking of the instruments. Experiments on a surgical phantom and in vivo porcine trials demonstrate that the proposed method supports multi-task operation, including compliant laparoscope manipulation and autonomous instrument tracking, while maintaining the RCM constraint and reducing sustained trocar-site loading.**

This work was supported by the National Natural Science Foundation of China under Grants 62322308, 62403178, 62133004, 72188101, and 72293585; by the Fundamental and Interdisciplinary Disciplines Breakthrough Plan of the Ministry of Education of China under Grant JYB2025XDXM109; by the Research Innovation Capacity Support Program for Young Teachers of Central Universities (SRICSPYF-ZY2025006); and by the Fundamental Research Funds for the Central Universities under Grant JZ2025HGTB0229. (Corresponding authors: Jin Fang.)

Xiaojian Li, Jin Fang, Yudong Shi, Xilin Xiao, Kang Min, Ling Li and Hangjie Mo are with the School of Management and the Key Laboratory of Process Optimization and Intelligent Decision-Making (Ministry of Education), Hefei University of Technology, Hefei 230009, China (email: fangjin@mail.hfut.edu.cn).

Kai Yan is with the Department of Thoracic Surgery, Shanghai Pulmonary Hospital, School of Medicine, Tongi University, Shanghai 200433, China.

Hua Tang is with the Department of Thoracic Surgery, Changhai Hospital, Shanghai 200433, China.

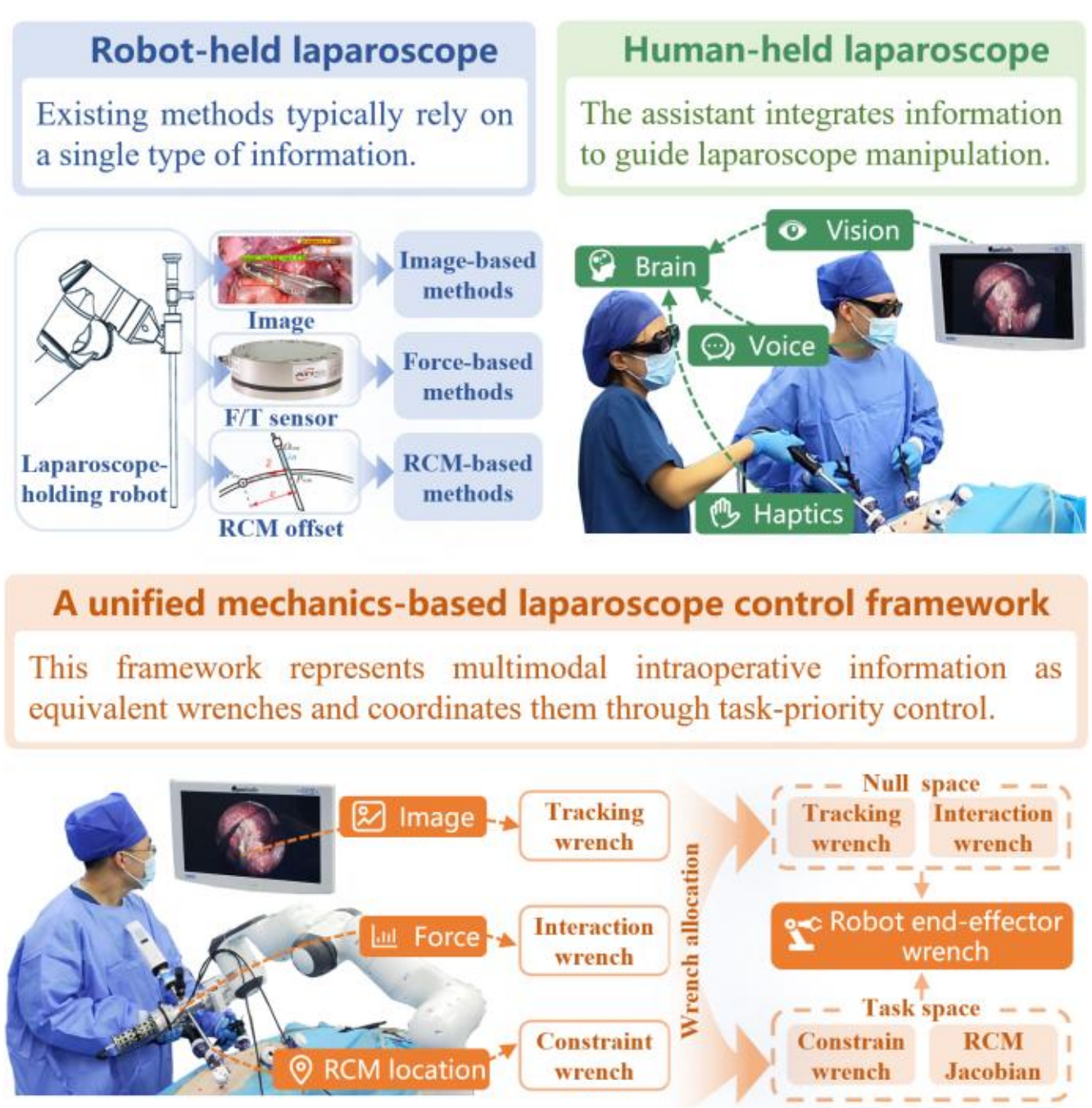

Figure 1. Overview of existing robot-held control, human-held manipulation, and the proposed unified mechanics-based laparoscope control framework.

## I. INTRODUCTION

MINIMALLY invasive surgery has reshaped modern surgical practice by reducing trauma, shortening hospital stays, and accelerating patient recovery [1], [2]. These benefits, however, come at the cost of demanding highly stable and well-controlled laparoscopic visualization in a confined workspace [3]. In conventional procedures, an assistant is responsible for manipulating the laparoscope, which can lead to view instability, limited reproducibility, and increased fatigue over the course of an operation [4]–[7]. Against this backdrop, laparoscope-holding robotic systems have been introduced to provide precise, repeatable, and surgeon-supportive laparoscope motion, and have gradually become an important branch of medical robotics in both research and clinical applications [4], [5], [8].

The central objective of laparoscope-holding robot control is to continuously maintain an appropriate intraoperative field of view (FOV) [4], [8]. As shown in Fig. 1, when the laparoscope is held by a human assistant, the assistant adjusts the laparoscope by integrating visual information about the surgical scene and instrument motion, auditory information about the surgeon's instructions and view requirements, and haptic information about the interaction state during laparoscope manipulation. By integrating information from these sources, the assistant continuously regulates the laparoscope pose to maintain an appropriate intraoperative view. In laparoscope-holding robotic systems, these cues can be explicitly acquired through dedicated sensors and perception algorithms. However, effectively integrating them remains challenging, because they differ markedly in physical meaning, units, and representation, making it difficult to build a unified framework for direct laparoscope motion generation.

Existing laparoscope-holding robot control methods are often developed around individual types of intraoperative information, making it difficult to fully exploit the diverse cues available during surgery. A large body of work has employed visual servoing to regulate the laparoscopic FOV autonomously [9]–[12]. In these methods, surgical instruments or task-relevant regions are extracted from the laparoscopic image and used as visual features for robot control. Existing studies have explored multi-objective optimization methods for simultaneous position and orientation regulation [13], control-barrier-function-based methods for autonomous endoscope motion [14], multi-instrument-tracking methods for maintaining regions of interest within the FOV [15], and deep-learning-based methods for learning laparoscope motion policies directly from surgical videos [16]–[18]. In these approaches, visual cues are used to generate viewpoint-adjustment commands, allowing the laparoscope to be regulated according to the image locations of instruments and regions of interest.

In addition to image-based regulation, physical interaction between the surgeon and the laparoscope has also been widely explored for camera control. Kim et al. utilized a six-axis force/torque sensor to realize direct teaching of a surgical assistant robot, allowing intuitive manual guidance while maintaining a virtual remote center of motion (RCM) [19]. Huang et al. integrated a load cell into a flexible endoscope holder and combined impedance control with software-defined RCM constraints, enabling compliant motion in response to surgeon-applied forces while preserving the insertion point and visual alignment [20]. Dede et al. further introduced a hybrid position-admittance control framework that maps forces measured at the endoscope to the pivot point, enabling smooth back drivable motion under RCM constraints [21]. In these studies, force information is used to translate surgeon-applied interaction forces into robot motion, thereby enabling intuitive guidance and adjustment of the laparoscope.

Another important direction concerns moving-RCM control, which addresses laparoscope manipulation when the trocar-related pivot point cannot be assumed fixed throughout the procedure. Early studies mainly extended kinematic formulations to accommodate nonstationary trocar points. Pham et al. proposed a moving RCM method based on an augmented Jacobian, enabling simultaneous regulation of trocar motion and end-effector kinematics [22]. Marinho et al. further developed a dual-quaternion-based constrained trajectory formulation that unified tool motion and trocar motion within a continuous representation and improved robustness against drift [23]. More recent studies have incorporated force feedback to achieve adaptive RCM behavior. Nasiri et al. introduced an admittance-based adaptive RCM framework that updates the RCM position according to estimated interaction forces [24]. Similarly, Fontúrbel et al. proposed a force-driven pivoting control strategy that minimizes trocar interaction forces without explicitly enforcing a fixed geometric RCM constraint [25]. In this line of work, RCM-related information is incorporated not merely to impose a fixed trocar constraint, but to continuously adapt the pivot condition of laparoscope motion according to trocar displacement or interaction forces.

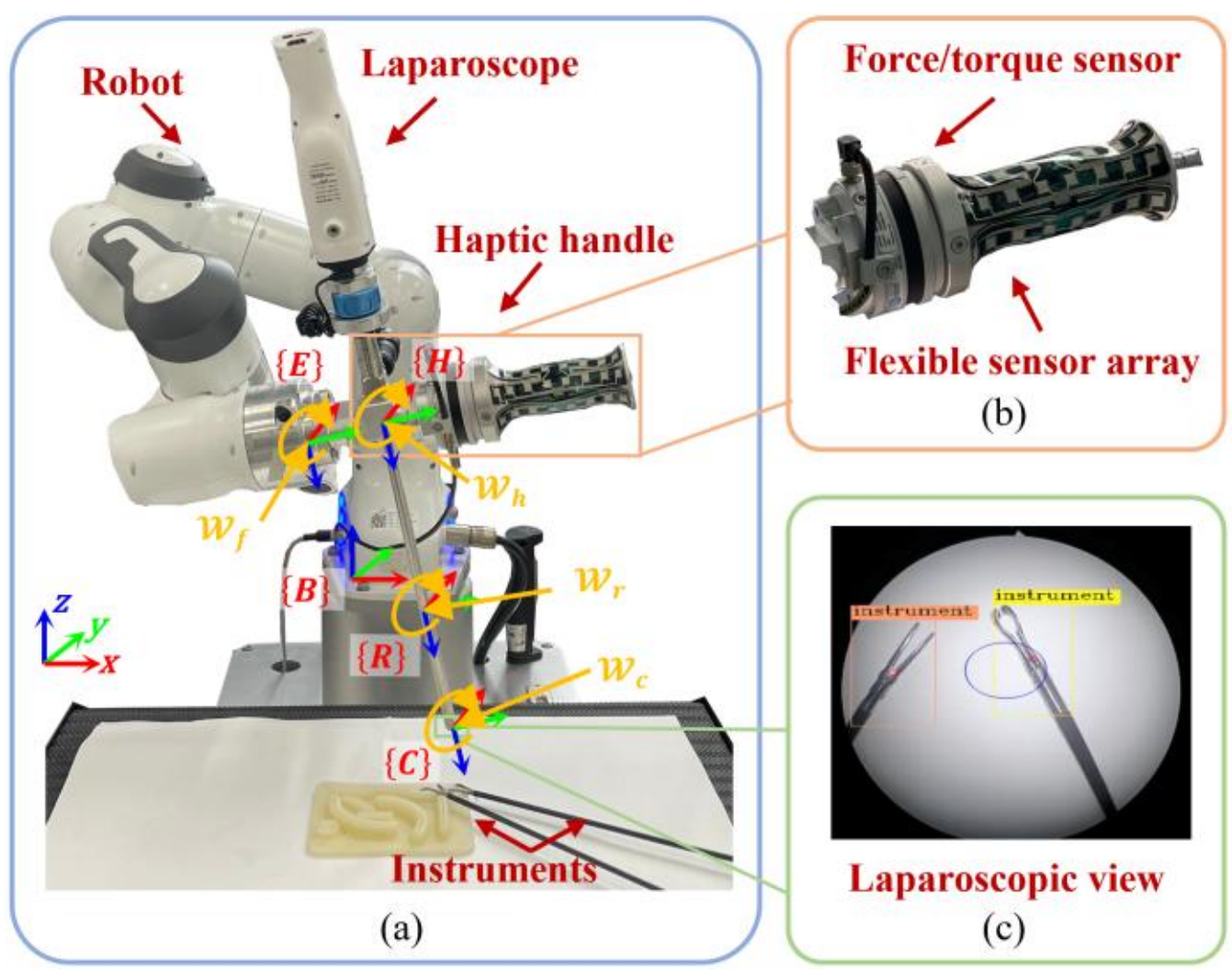

Figure 2. System overview of the proposed laparoscope-holding robot. (a) Overall experimental platform with the robot, laparoscope, sensorized haptic handle, defined coordinate frames, and representative wrenches. (b) Sensorized haptic handle integrating a six-axis force/torque sensor and a flexible sensor array. (c) Representative laparoscopic view with detected surgical instruments.

Overall, prior work has thoroughly explored the use of individual intraoperative information channels for laparoscope control. Even when multiple modalities have been considered jointly, their integration is often achieved in a task-specific manner rather than through a unified control-level formulation. For example, Fang et al. [26] attempted to combine force-driven and vision-driven cues for autonomous laparoscope control. However, the method was still developed as a hybrid strategy tailored to these two specific information sources, without establishing a common control-level representation for systematically incorporating additional heterogeneous intraoperative cues. As a result, visual observations, interaction forces, and RCM-related variables are typically expressed in different forms and play different roles in motion generation, which makes their direct coordination within a single control process difficult. To address this issue, a key requirement is to establish a common physical representation through which heterogeneous intraoperative cues can be expressed, coordinated, and directly converted into laparoscope motion commands.

Motivated by this need, this work proposes a unified mechanics-based control framework for laparoscope-holding robots to achieve consistent representation and cooperative fusion of multimodal intraoperative information. Specifically, position, force/torque, and image cues are mapped into an equivalent-wrench representation in the operational space, providing a common physical form for control-oriented multimodal fusion. Based on this representation, a task-priority scheme is developed to inject different wrenches into the task space and null space, thereby synthesizing laparoscope control commands while coordinating multiple objectives within a single framework. As representative examples, we construct an RCM-constraint wrench to enforce the geometric RCM constraint and reduce trocar interaction force, a laparoscope-manipulation wrench to support compliant dragging, and an instrument-tracking wrench to realize autonomous visual tracking of surgical instruments. Experimental evaluation on a surgical phantom and in vivo animal models verifies that the proposed framework can support compliant laparoscope manipulation and autonomous instrument tracking while maintaining the RCM constraint, demonstrating its effectiveness for multimodal laparoscope control in robotic surgery.

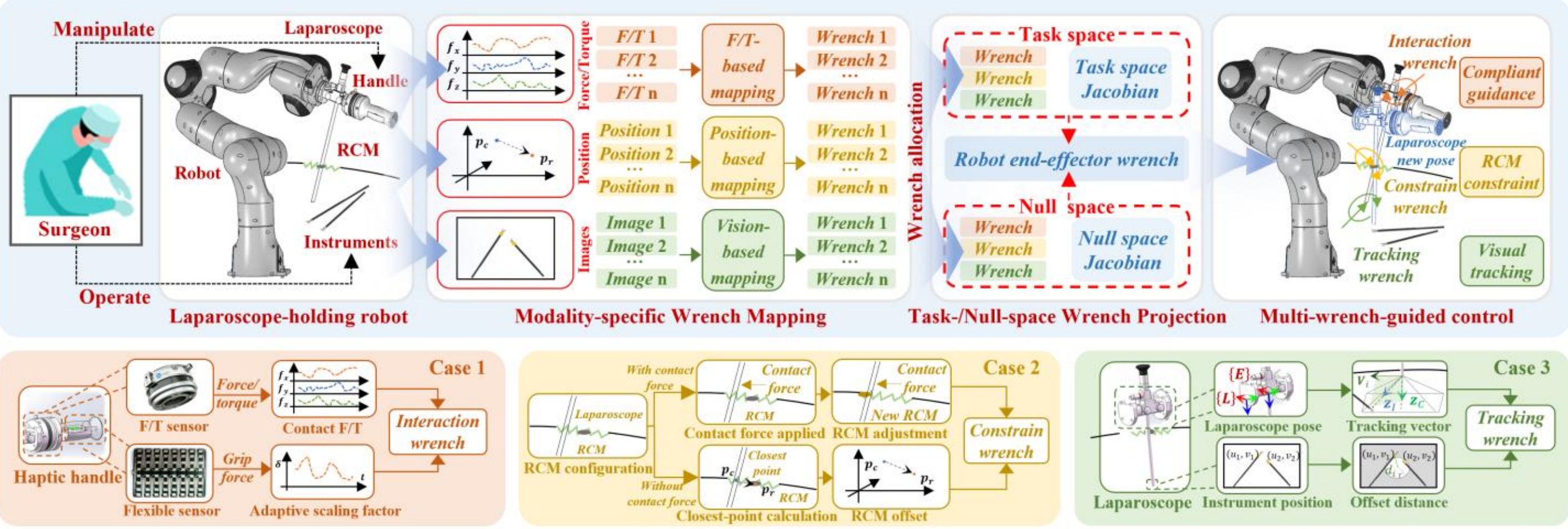

Figure 3. The framework of the proposed laparoscope-holding robot.

## II. Laparoscope-Holding Robot System

### A. System Description

The overall architecture of the proposed laparoscope-holding robot is shown in Fig. 2. The platform consists of three main hardware components: a 30° laparoscope, a handle with integrated force and tactile sensing, and a 7-DoF collaborative robot.

The laparoscope (J1830, Shenyang Shenda Laparoscope Co., Ltd., China) is rigidly mounted to the end-effector of the robot and provides real-time images of the intra-abdominal scene. A YOLOv7-based detector is employed to identify and localize surgical instruments in the laparoscopic view. The detected instrument bounding boxes are converted into desired robot motion directions within the vision-driven part of the control scheme, thereby enabling automatic instrument-centered FOV regulation.

To support intuitive physical interaction between the surgeon and the robot, a sensorized tactile handle is attached to the laparoscope. The handle integrates a six-axis force/torque sensor (HEX-E-V2, OnRobot, Denmark) and a flexible sensor array (Zhongke Benyuan Information Technology Co., Ltd., China). The force/torque sensor measures the interaction wrench applied by the surgeon to the laparoscope and provides the force-driven input for hand-guided motion. The flexible sensor array is used to estimate the grip tightness on the handle. It consists of 64 sensing elements, each outputting a nominal value of 0 in the absence of contact and a value in the range (0, 255] when loaded, with larger values corresponding to higher local contact forces on the handle surface. The interaction wrench and grip information obtained from the tactile handle are used as inputs to the force-driven component of the proposed scheme.

The collaborative robot employed in this system is a 7-DoF Franka Emika arm (Franka Emika GmbH, Germany). Each joint of the Franka robot is equipped with an integrated torque sensor that provides real-time measurements of joint torques. The measured joint torques are mapped via the robot Jacobian to an equivalent Cartesian wrench at the end-effector, from which the forces and torques acting on the laparoscope and at the RCM are estimated. The resulting estimated external wrenches are then used to adaptively update the position of the RCM.

### B. Definition of Coordinate Frames

To describe the kinematic relationships of the laparoscope-holding robotic system, several coordinate frames are introduced. As shown in Fig. 2(a), the robot base frame $\{B\}$ is fixed at the robot base and serves as the global reference frame of the system. The end-effector frame $\{E\}$ is attached to the robot end-effector at the location where the laparoscope is mounted. In addition, a handle frame $\{H\}$, an RCM frame $\{R\}$, and a camera frame $\{C\}$ are defined to represent the laparoscope handle, the RCM, and the camera center, respectively. Unless otherwise specified, the axes of $\{H\}$, $\{R\}$, and $\{C\}$ are chosen to be parallel to those of $\{E\}$, such that these frames differ from $\{E\}$ only in the locations of their origins.

### C. Overall Framework

The overall system framework of the proposed unified mechanics-based control method is illustrated in Fig. 3. During surgery, while the lead surgeon operates the instruments and interacts with the laparoscope-holding robot, the robotic system continuously acquires multimodal intraoperative information, including force/torque, position, and image cues. These heterogeneous signals are first converted into equivalent wrenches through modality-specific mappings, so that they can be represented in a common mechanical form in the operational space. The resulting wrenches are then allocated through a hierarchical task-priority structure. Specifically, the RCM-related wrench is projected into the task space through the task-space Jacobian to enforce the primary geometric constraint of laparoscope motion, whereas the interaction wrench and the tracking wrench are projected into the null space through the null-space Jacobian to realize lower-priority objectives without violating the RCM requirement. Through this task/null-space wrench allocation, the multimodal cues are fused into a unified robot end-effector wrench, from which the final laparoscope motion is generated. Within this framework, three representative wrench designs are constructed: the RCM-constraint wrench is projected into the task space through the task-space Jacobian to enforce the primary geometric constraint of laparoscope motion and reduce trocar-related contact loading, whereas the interaction wrench and the tracking wrench are projected into the null space through the null-space Jacobian to realize compliant laparoscope guidance and autonomous instrument-centered FOV regulation without violating the RCM requirement.

## III. Method

This section describes the proposed control method from the perspective of unified mechanics modeling. To provide a common control-oriented representation for heterogeneous intraoperative information, the method first formulates the robot dynamics and the corresponding torque-level redundancy-resolution framework under the RCM constraint. On this basis, interaction, vision, and RCM-related cues are respectively converted into wrenches. Finally, these wrenches are distributed between the task space and the null space, enabling coordinated execution of RCM maintenance, compliant laparoscope manipulation, and autonomous instrument tracking within a single control framework.

### *A. Robot Dynamics and Torque-Level Task-Priority Control*

This subsection introduces the dynamic model of the redundant manipulator and establishes the torque-level task-priority control formulation. By decomposing the control torque into an RCM-related task component and a null-space component, the framework provides the basis for enforcing the primary RCM constraint while preserving redundancy for lower-priority objectives.

The dynamic model of a $n$-DOF serial manipulator in joint-space can be expressed as:

$$\boldsymbol{M}(\boldsymbol{\theta})\ddot{\boldsymbol{\theta}} + \boldsymbol{C}(\boldsymbol{\theta},\dot{\boldsymbol{\theta}})\dot{\boldsymbol{\theta}} + \boldsymbol{G}(\boldsymbol{\theta}) = \boldsymbol{\tau}, \tag{1}$$

where the joint vector is denoted by $\boldsymbol{\theta} \in \mathbb{R}^n$, the inertia matrix by $\boldsymbol{M}(\boldsymbol{\theta}) \in \mathbb{R}^{n\times n}$ , the centrifugal and Coriolis term by $\boldsymbol{C}(\boldsymbol{\theta},\dot{\boldsymbol{\theta}})\dot{\boldsymbol{\theta}} \in \mathbb{R}^n$, the gravitational term by $\boldsymbol{G}(\boldsymbol{\theta}) \in \mathbb{R}^n$, and the control torque by $\boldsymbol{\tau}$.

To realize torque-level task prioritization under the RCM constraint, the control torque is decomposed into

$$\boldsymbol{\tau}(\boldsymbol{\theta}) = \boldsymbol{J}_R{}^T(\boldsymbol{\theta})\boldsymbol{F}_r + \boldsymbol{N}_r{}^T(\boldsymbol{\theta})\boldsymbol{\tau}_n, \tag{2}$$

where the RCM task Jacobian is denoted by $\boldsymbol{J}_r \in \mathbb{R}^{3\times n}$, which maps joint velocities to the task-space rate of the RCM constraint error; the generalized task-space force produced by the RCM controller is denoted by $\boldsymbol{F}_r$; and the auxiliary joint-torque command for lower-priority objectives is denoted by $\boldsymbol{\tau}_n$. The null-space projector of the RCM task $\boldsymbol{N}_r$ is given by

$$\boldsymbol{N}_r(\boldsymbol{\theta}) = \boldsymbol{I}_n - \boldsymbol{J}_r{}^+(\boldsymbol{\theta})\boldsymbol{J}_r(\boldsymbol{\theta}), \tag{3}$$

where the identity matrix is denoted by $\boldsymbol{I}_n \in \mathbb{R}^{n\times n}$, and the mass-matrix–weighted pseudo-inverse $\boldsymbol{J}_r{}^+(\boldsymbol{\theta})$ is computed as follows

$$\boldsymbol{J}_r{}^+(\boldsymbol{\theta}) = \boldsymbol{M}^{-1}(\boldsymbol{\theta})\boldsymbol{J}_r{}^T(\boldsymbol{\theta})\left(\boldsymbol{J}_r(\boldsymbol{\theta})\boldsymbol{M}^{-1}(\boldsymbol{\theta})\boldsymbol{J}_r{}^T(\boldsymbol{\theta})\right)^{-1}. \tag{4}$$

Therefore, the control torque is decomposed into two components: the torque generated for enforcing the RCM constraint and the auxiliary torque projected into the null space of the RCM task. With this formulation, the primary RCM task is guaranteed to be satisfied, while lower-priority objectives can be executed without affecting the fulfillment of the RCM constraint.

### *B. Wrench Formulation from Multimodal Information*

This subsection presents the design of the wrenches used in the proposed framework. To express heterogeneous intraoperative information in a unified mechanical form, surgeon interaction, image-based guidance, and RCM regulation are respectively formulated as equivalent wrenches, which can then be incorporated into the subsequent hierarchical control scheme.

*1) The interaction wrench $\boldsymbol{\mathcal{W}}_h$:* The interaction wrench is determined by the surgeon-applied wrench measured by the force/torque sensor and the grip force estimated from the flexible sensor array. The interaction wrench incorporates an adaptive gain mechanism to prevent inadvertent activation caused by incidental contact.

The surgeon-operated wrench acting on the force/torque sensor is represented by

$$\boldsymbol{\mathcal{W}}_s = \begin{bmatrix}\boldsymbol{M}_s\\ \boldsymbol{F}_s\end{bmatrix} = \begin{bmatrix}M_s^x\boldsymbol{x}_H + M_s^y\boldsymbol{y}_H + M_s^z\boldsymbol{z}_H\\ F_s^x\boldsymbol{x}_H + F_s^y\boldsymbol{y}_H + F_s^z\boldsymbol{z}_H\end{bmatrix}, \tag{5}$$

where the surgeon-operated wrench is denoted by $\boldsymbol{\mathcal{W}}_s \in \mathbb{R}^6$, the force and torque readings of the force/torque sensor by $\boldsymbol{M}_s \in \mathbb{R}^3$ and $\boldsymbol{F}_s \in \mathbb{R}^3$, respectively, and the unit basis vectors of frame $\{H\}$ expressed in the base frame $\{B\}$ by $\boldsymbol{x}_H$, $\boldsymbol{y}_H$, and $\boldsymbol{z}_H$.

Under the RCM constraint, the laparoscope cannot undergo arbitrary 6-DOF rigid-body motion. Instead, its admissible motion is reduced to four independent DOFs, namely, two rotations about the RCM, one translation along the laparoscope shaft, and one axial rotation about the shaft. Therefore, only the three force components and the axial torque component are retained to construct the interaction wrench. The interaction wrench $\boldsymbol{\mathcal{W}}_h$ can be computed as follows:

$$\boldsymbol{\mathcal{W}}_h = \begin{bmatrix}\boldsymbol{M}_h\\ \boldsymbol{F}_h\end{bmatrix} = \delta\begin{bmatrix}M_s^z\boldsymbol{z}_H\\ F_s^x\boldsymbol{x}_H + F_s^y\boldsymbol{y}_H + F_s^z\boldsymbol{z}_H\end{bmatrix}, \tag{6}$$

where the interaction wrench is denoted by $\boldsymbol{\mathcal{W}}_h \in \mathbb{R}^6$, and its torque and force components are denoted by $\boldsymbol{M}_h \in \mathbb{R}^3$ and $\boldsymbol{F}_h \in \mathbb{R}^3$, respectively.

The parameter $\delta$ denotes the gain of the interaction wrench and is determined from the tactile sensor array as

$$\delta = \begin{cases} 0, & n_c \le k_{min}, \\ \dfrac{\sum_{j=1}^{64} Se_j}{k_H}, & n_c > k_{min}. \end{cases} \tag{7}$$

where $Se_j$ denotes the measurement of the $j$-th sensing element, $n_c$ denotes the number of sensing elements under contact, and $k_{min}$ denotes the minimum number of contacted sensing elements required to activate the interaction control. The parameter $k_H$ is a normalization factor used to scale the summed tactile response to a reasonable range. In this way, the interaction wrench is enabled only when the handle is grasped with a sufficient effective contact area, as approximated by the number of activated sensing elements, thereby reducing the risk of unintended laparoscope motion caused by incidental touch.

*2) The virtual wrench $\boldsymbol{\mathcal{W}}_c$:* The virtual wrench, denoted by $\boldsymbol{\mathcal{W}}_c \in \mathbb{R}^6$, is introduced to prevent surgical instruments from moving out of the FOV. It can be computed as follows:

$$\boldsymbol{\mathcal{W}}_c = \begin{bmatrix} \boldsymbol{M}_c \\ \boldsymbol{F}_c \end{bmatrix} = \begin{cases} \boldsymbol{0}_{6\times 1}, & n_s = 0, \\ \begin{bmatrix} \boldsymbol{0}_{3\times 1} \\ \sum_{i=1}^{n_s} F_v(d_i)\boldsymbol{V}_i \end{bmatrix}, & n_s > 0, \end{cases} \tag{8}$$

where the torque and force components of the virtual wrench are denoted by $\boldsymbol{M}_c \in \mathbb{R}^3$ and $\boldsymbol{F}_c \in \mathbb{R}^3$, respectively. The variable $n_s$ denotes the number of surgical instruments in the image. The distance $d_i$ is computed from the image coordinates $(u_i, v_i)$ as

$$d_i = \sqrt{(u_i - u_0)^2 + (v_i - v_0)^2}, \tag{9}$$

where the principal point is denoted by $(u_0, v_0)$. The magnitude profile $F_v(d_i) \in [0, F_{max}]$ is defined by the following piecewise function:

$$F_v(d_i) = \begin{cases} 0, & 0 \le d_i \le \epsilon_1, \\ \dfrac{F_{max}}{2}\left(\sin\dfrac{(d_i - \epsilon_1)\pi}{2(\epsilon_2 - \epsilon_1)} + 1\right), & \epsilon_1 < d_i \le \epsilon_2, \\ F_{max}, & \epsilon_2 < d_i, \end{cases} \tag{10}$$

where the dead-zone radius around the image center is denoted by $\epsilon_1$, the outer activation radius by $\epsilon_2$, and the maximum achievable force magnitude by $F_{max}$.

The direction $\boldsymbol{V}_i$ is computed using the unit-depth ray $\boldsymbol{\mathcal{P}}'_i$ in the laparoscope frame:

$$\boldsymbol{\mathcal{P}}'_i = \frac{u_i - u_0}{f_x} \cdot \boldsymbol{x}_C + \frac{v_i - v_0}{f_y} \cdot \boldsymbol{y}_C + \boldsymbol{z}_C, \tag{11}$$

where the camera focal lengths in pixel units are denoted by $f_x$ and $f_y$, and the unit basis vectors of frame $\{C\}$ expressed in the base frame $\{B\}$ are denoted by $\boldsymbol{x}_C$, $\boldsymbol{y}_C$, and $\boldsymbol{z}_C$. The in-plane direction is obtained by projecting $\boldsymbol{\mathcal{P}}'_i$ onto the image plane:

$$\boldsymbol{V}_i = \frac{(\boldsymbol{z}_C \times \boldsymbol{\mathcal{P}}'_i) \times \boldsymbol{\mathcal{P}}'_i}{\left\| (\boldsymbol{z}_C \times \boldsymbol{\mathcal{P}}'_i) \times \boldsymbol{\mathcal{P}}'_i \right\|}. \tag{12}$$

This construction yields a continuous corrective action that remains inactive near the image center and smoothly saturates to $F_{max}$ as the instrument approaches the FOV boundary.

*3) The RCM constraint wrench $\boldsymbol{\mathcal{W}}_r$:* To enforce the RCM constraint during laparoscope manipulation, an RCM constraint wrench $\boldsymbol{\mathcal{W}}_r \in \mathbb{R}^6$ is introduced. Since the RCM regulation considered here is formulated in terms of the positional deviation of the constraint point, the corrective action is constructed as a translational force, and the corresponding wrench is written as

$$\boldsymbol{\mathcal{W}}_r \begin{bmatrix} \boldsymbol{M}_r \\ \boldsymbol{F}_r \end{bmatrix} = \begin{bmatrix} \boldsymbol{0}_{3\times 1} \\ \boldsymbol{F}_r \end{bmatrix}, \tag{13}$$

where

$$\boldsymbol{F}_r = -\boldsymbol{K}_p \boldsymbol{e}_R(\boldsymbol{\theta}) - \boldsymbol{K}_d \dot{\boldsymbol{e}}_R(\boldsymbol{\theta}). \tag{14}$$

The positive-definite gain matrices are denoted by $\boldsymbol{K}_p \in \mathbb{R}^{3\times 3}$ and $\boldsymbol{K}_d \in \mathbb{R}^{3\times 3}$, and the RCM error by $\boldsymbol{e}_R(\boldsymbol{\theta}) \in \mathbb{R}^3$, whose geometric definition is given later.

### C. Wrench Allocation in Task Space and Null Space

This subsection describes how the designed wrenches are incorporated into the hierarchical controller. Specifically, the RCM-related wrench is allocated in the task space to enforce the primary geometric constraint, whereas the interaction, virtual, and damping wrenches are introduced in the null space to realize compliant manipulation and instrument-guided viewpoint adjustment without violating the RCM requirement.

*1) Task-Space Formulation of the RCM Constraint:* To formulate the primary RCM task, this part establishes the geometric quantities required in the task space. Specifically, a dynamic update mechanism is introduced for the RCM point, after which the RCM error and the corresponding Jacobian are defined for subsequent control construction.

The forward kinematics of the end-effector frame $\{E\}$ with respect to the base frame $\{B\}$ is given by

$$\boldsymbol{T}_E^B(\boldsymbol{\theta}) = \begin{bmatrix} \boldsymbol{R}_E^B(\boldsymbol{\theta}) & \boldsymbol{p}_E^B(\boldsymbol{\theta}) \\ \boldsymbol{0}_{1\times 3} & 1 \end{bmatrix}, \tag{15}$$

where $\boldsymbol{R}_E^B(\boldsymbol{\theta}) \in SO(3)$ is the rotation matrix of frame $\{E\}$ relative to frame $\{B\}$, and $\boldsymbol{p}_E^B(\boldsymbol{\theta}) \in \mathbb{R}^3$ is the position vector of the origin of frame $\{E\}$ expressed in frame $\{B\}$. For notational simplicity, the superscript $\{B\}$ indicating quantities expressed in the base frame is omitted in the remainder of this paper unless otherwise stated.

A laparoscope of length $L$ is aligned with the end-effector $z$-axis. Denoting $\boldsymbol{I}_z = [0 \quad 0 \quad 1]^T$, the laparoscope direction expressed in the base frame is defined as

$$\boldsymbol{d}_z(\boldsymbol{\theta}) = \boldsymbol{R}_E(\boldsymbol{\theta})\boldsymbol{I}_z, \ \ \|\boldsymbol{d}_z(\boldsymbol{\theta})\| = 1. \tag{16}$$

The projection matrix onto the plane orthogonal to the laparoscope axis is defined as

$$\boldsymbol{P}(\boldsymbol{\theta}) = \boldsymbol{I}_3 - \boldsymbol{d}_z(\boldsymbol{\theta})\boldsymbol{d}_z{}^T(\boldsymbol{\theta}), \tag{17}$$

At the initialization instant $t_0$, the initial RCM point $\boldsymbol{p}_R(t_0) \in \mathbb{R}^3$ is selected on the laparoscope axis as

$$\boldsymbol{p}_R(t_0) = \boldsymbol{p}_E(\boldsymbol{\theta}_0) + \lambda_0 L \boldsymbol{d}_z(\boldsymbol{\theta}_0), \tag{18}$$

where $\lambda_0$ specifies the location of the RCM point along the laparoscope at the initial configuration $\boldsymbol{\theta}_0$.

During surgery, however, the trocar-related pivot location may vary because of soft-tissue deformation and interaction between the laparoscope and the incision site. To account for this effect, the RCM point is allowed to adapt online according to a residual-force signal constructed from the measured and estimated interaction forces. Specifically, the difference between the force/torque sensor measurement and the robot-estimated end-effector force is defined as $\Delta\boldsymbol{F} = [\Delta F^x \quad \Delta F^y \quad \Delta F^z]^T \in \mathbb{R}^3$. Taking the $x$-axis component as an example, let

$$\Delta F^x = F_s^x - F_m^x, \tag{19}$$

where $F_s^x$ is the $x$-axis force component measured by the six-axis force sensor wrench $\boldsymbol{\mathcal{W}}_s$, and $F_m^x$ is the corresponding $x$-axis force component of the robot-estimated end-effector wrench $\boldsymbol{\mathcal{W}}_m$. To improve robustness, the residual force is passed through a dead-zone–saturation nonlinearity. The mapped correction force is defined as

$$F_{re}^x(\Delta F^x) = sgn(\Delta F^x)\Phi(|\Delta F^x|), \tag{20}$$

where the sign function $sgn(\cdot)$ restores the direction of the residual force after the magnitude-based nonlinear mapping. The scalar mapping function $\Phi(\cdot)$ is defined as

$$\Phi(|\Delta F^x|) = \begin{cases} 0, & 0 \le |\Delta F^x| \le \gamma_1, \\ \dfrac{F_{op}}{2}(\sin\eta_x + 1), & \gamma_1 < |\Delta F^x| \le \gamma_2, \\ F_{op}, & \gamma_2 < |\Delta F^x|, \end{cases} \tag{21}$$

with

$$\eta_x = \frac{(|\Delta F^x| - \gamma_1)\pi}{2(\gamma_2 - \gamma_1)}. \tag{22}$$

Here, the parameter $\gamma_1$ specifies the dead-zone threshold below which the residual force is ignored. The parameter $\gamma_2$ specifies the upper bound of the smooth transition region. The quantity $F_{op}$ denotes the saturation value of the mapped correction force. In this way, small residual forces are suppressed to improve robustness against noise, medium residual forces are smoothly amplified to avoid discontinuous corrections, and large residual forces are bounded to prevent excessive adjustment. The same mapping is applied to the y-axis and z-axis components to obtain the residual-force vector $\boldsymbol{F}_{re}$.

Since shifting the RCM point is mainly a lateral correction, along-shaft force is more related to insertion friction and should not drive $\boldsymbol{p}_R$ drift. Define the lateral contact force component:

$$\boldsymbol{F}_{re}^{\perp} = \boldsymbol{P}(\boldsymbol{\theta})\boldsymbol{F}_{re}. \tag{23}$$

Therefore, only the lateral component $\boldsymbol{F}_{re}^{\perp}$ of the residual contact force is used to adapt the RCM point. Let $\boldsymbol{p}_R(t) \in \mathbb{R}^3$ denote the adaptive RCM point at time $t$, and let $\boldsymbol{p}_R^*(t) \in \mathbb{R}^3$ denote a reference RCM point used as an anchor for the adaptation. The time evolution of $\boldsymbol{p}_R(t)$ is governed by

$$\dot{\boldsymbol{p}}_R(t) = -\alpha \boldsymbol{F}_{re}^{\perp} - \beta\big(\boldsymbol{p}_R(t) - \boldsymbol{p}_R^*(t)\big), \tag{24}$$

where $\dot{\boldsymbol{p}}_R(t)$ is the time derivative of $\boldsymbol{p}_R(t)$, $\alpha$ is the adaptation compliance, and $\beta$ is the leakage coefficient. The first term drives the motion of the adaptive RCM point according to the lateral residual contact force, whereas the second term weakly attracts $\boldsymbol{p}_R(t)$ toward the reference point $\boldsymbol{p}_R^*(t)$ to prevent unbounded drift.

To avoid frequent redefinition of the reference point during transient motion, the reference RCM point $\boldsymbol{p}_R^*(t)$ is modeled as a piecewise constant function. Specifically, the update instants are denoted by $\{t_k\}_{k=0}^{\infty}$, where $t_k$ is the $k$-th update instant, and they are determined by

$$\|\dot{\boldsymbol{p}}_R(t_k)\| \le \varepsilon, \tag{25}$$

where the prescribed threshold is denoted by $\varepsilon$. At each update instant $t_k$, the reference point is reset to the current adaptive RCM point, that is,

$$\boldsymbol{p}_R^*(t) = \boldsymbol{p}_R(t_k), \quad t \in [t_k, t_{k+1}], \tag{26}$$

which means that $\boldsymbol{p}_R^*(t)$ remains constant between two consecutive update instants and is replaced by the current adapted RCM point only after $\boldsymbol{p}_R(t)$ has approximately converged to a new location.

Once the velocity $\dot{\boldsymbol{p}}_R(t)$ is determined, the adaptive RCM point $\boldsymbol{p}_R(t)$ is updated by time integration of $\dot{\boldsymbol{p}}_R(t)$. Based on the current adaptive RCM point $\boldsymbol{p}_R(t)$, the RCM error is defined as the lateral deviation of the laparoscope axis from $\boldsymbol{p}_R(t)$, and the corresponding RCM error is given by

$$\boldsymbol{e}_R(\boldsymbol{\theta}, t) = \boldsymbol{P}(\boldsymbol{\theta})\big(\boldsymbol{p}_R(t) - \boldsymbol{p}_E(\boldsymbol{\theta})\big). \tag{27}$$

where the component of $\boldsymbol{p}_R(t) - \boldsymbol{p}_E(\boldsymbol{\theta})$ perpendicular to the laparoscope axis is denoted by $\boldsymbol{e}_R(\boldsymbol{\theta}, t)$. Accordingly, the error varies with both the robot configuration $\boldsymbol{\theta}$ and the adaptive evolution of $\boldsymbol{p}_R(t)$. The corresponding Jacobian is

$$\boldsymbol{J}_R(\boldsymbol{\theta}, t) = \frac{\partial \boldsymbol{e}_R(\boldsymbol{\theta}, t)}{\partial \boldsymbol{\theta}} \in \mathbb{R}^{3\times n}. \tag{28}$$

This Jacobian is subsequently used together with the RCM constraint wrench defined in Section III-B3 to generate the task-space control torque for the primary RCM task.

*2) Null-Space Mapping via the Closest-Point Jacobian:* To realize the lower-priority objectives in the null space, the geometric quantities required for wrench-to-torque mapping are next established. Specifically, the closest point on the laparoscope shaft is introduced as a common reference point, and the corresponding geometric Jacobian is derived for subsequent null-space control construction.

The geometric Jacobian of the closest point on the laparoscope shaft is derived in this subsection. This Jacobian is subsequently used to map task-space wrenches into joint torques.

The scalar projection of $\boldsymbol{p}_R$ onto the laparoscope axis is

$$s(\boldsymbol{\theta}, t) = \frac{\boldsymbol{d}_z(\boldsymbol{\theta})^T\big(\boldsymbol{p}_R(t) - \boldsymbol{p}_E(\boldsymbol{\theta})\big)}{L}. \tag{29}$$

Since the laparoscope is a finite segment, we clamp the parameter to [0,1]:

$$s^*(\boldsymbol{\theta},t) = arg \min_{\psi\in[0,1]} |\psi - s(\boldsymbol{\theta},t)|. \tag{30}$$

The closest point $\boldsymbol{p}_G(\boldsymbol{\theta})$ and its offset $\boldsymbol{o}_G(\boldsymbol{\theta})$ from the end-effector origin are then

$$\boldsymbol{p}_G(\boldsymbol{\theta},t) = \boldsymbol{p}_E(\boldsymbol{\theta}) + s^*(\boldsymbol{\theta},t)L\boldsymbol{d}_z(\boldsymbol{\theta}), \tag{31}$$

$$\boldsymbol{o}_G(\boldsymbol{\theta},t) = \boldsymbol{p}_G(\boldsymbol{\theta},t) - \boldsymbol{p}_E(\boldsymbol{\theta}) = s^*(\boldsymbol{\theta},t)L\boldsymbol{d}_z(\boldsymbol{\theta}). \tag{32}$$

Let the geometric Jacobian at the end-effector origin be

$$\begin{bmatrix}\dot{\boldsymbol{\varphi}}_E(\boldsymbol{\theta})\\ \dot{\boldsymbol{p}}_E(\boldsymbol{\theta})\end{bmatrix} = \boldsymbol{J}_E\dot{\boldsymbol{\theta}} = \begin{bmatrix}\boldsymbol{J}_{E_\omega}(\boldsymbol{\theta})\\ \boldsymbol{J}_{E_v}(\boldsymbol{\theta})\end{bmatrix}\dot{\boldsymbol{\theta}}, \tag{33}$$

where the joint velocity vector is denoted by $\dot{\boldsymbol{\theta}} \in \mathbb{R}^n$, and the geometric Jacobian of the end-effector by $\boldsymbol{J}_E \in \mathbb{R}^{6\times n}$. Specifically, $\boldsymbol{J}_{E_\omega}(\boldsymbol{\theta}) \in \mathbb{R}^{3\times n}$ maps $\dot{\boldsymbol{\theta}}$ to the end-effector angular velocity $\dot{\boldsymbol{\varphi}}_E(\boldsymbol{\theta}) \in \mathbb{R}^3$, whereas $\boldsymbol{J}_{E_v}(\boldsymbol{\theta}) \in \mathbb{R}^{3\times n}$ maps $\dot{\boldsymbol{\theta}}$ to the linear velocity of the end-effector origin $\dot{\boldsymbol{p}}_E(\boldsymbol{\theta}) \in \mathbb{R}^3$, with both velocities expressed in the base frame $\{B\}$.

In the Jacobian construction, $s^*(\boldsymbol{\theta},t)$ is treated as an instantaneous scalar parameter frozen at the current configuration. For a point attached to the end-effector with offset $\boldsymbol{o}_g(\boldsymbol{\theta},t)$, its linear velocity satisfies

$$\dot{\boldsymbol{p}}_G(\boldsymbol{\theta},t) \approx \dot{\boldsymbol{p}}_E(\boldsymbol{\theta}) + \dot{\boldsymbol{\varphi}}_E(\boldsymbol{\theta}) \times \boldsymbol{o}_G(\boldsymbol{\theta},t). \tag{34}$$

Using the skew-symmetric matrix $[\boldsymbol{o}_G(\boldsymbol{\theta},t)]_\times$ such that $\dot{\boldsymbol{\varphi}}_E(\boldsymbol{\theta}) \times \boldsymbol{o}_G(\boldsymbol{\theta},t) = -[\boldsymbol{o}_G(\boldsymbol{\theta},t)]_\times \boldsymbol{J}_{E_\omega}(\boldsymbol{\theta})\dot{\boldsymbol{\theta}}$, the linear-velocity Jacobian for the closest point is

$$\boldsymbol{J}_{G_v}(\boldsymbol{\theta},t) = \boldsymbol{J}_{E_v}(\boldsymbol{\theta}) - [\boldsymbol{o}_G(\boldsymbol{\theta},t)]_\times \boldsymbol{J}_{E_\omega}(\boldsymbol{\theta}), \tag{35}$$

and the corresponding 6D geometric Jacobian is

$$\boldsymbol{J}_G(\boldsymbol{\theta},t) = \begin{bmatrix}\boldsymbol{J}_{G_\omega}(\boldsymbol{\theta})\\ \boldsymbol{J}_{G_v}(\boldsymbol{\theta},t)\end{bmatrix} \in \mathbb{R}^{6\times n}. \tag{36}$$

Since the point is rigidly attached to the laparoscope shaft, its angular-velocity Jacobian is identical to that of the end-effector, i.e., $\boldsymbol{J}_{G_\omega}(\boldsymbol{\theta}) = \boldsymbol{J}_{E_\omega}(\boldsymbol{\theta})$.

The corresponding joint-space torque is obtained via the dual Jacobian mapping:

$$\boldsymbol{\tau}_n = \boldsymbol{J}_G^T(\boldsymbol{\theta},t)\boldsymbol{W}_n. \tag{37}$$

The null-space wrench vector $\boldsymbol{W}_n$ expressed about the closest point is defined as:

$$\boldsymbol{W}_n = \boldsymbol{Q}_h\boldsymbol{W}_h + \boldsymbol{Q}_c\boldsymbol{W}_c + \boldsymbol{W}_f, \tag{38}$$

where

$$\boldsymbol{Q}_h = \begin{bmatrix}\boldsymbol{I}_3 & [\boldsymbol{p}_H - \boldsymbol{p}_G]_\times\\ \boldsymbol{0}_{3\times3} & \boldsymbol{I}_3\end{bmatrix}, \boldsymbol{Q}_c = \begin{bmatrix}\boldsymbol{I}_3 & [\boldsymbol{p}_C - \boldsymbol{p}_G]_\times\\ \boldsymbol{0}_{3\times3} & \boldsymbol{I}_3\end{bmatrix}. \tag{39}$$

Here, $\boldsymbol{Q}_h$ and $\boldsymbol{Q}_c$ are wrench transformation matrices that rewrite the wrenches defined about points $\boldsymbol{p}_H$ and $\boldsymbol{p}_C$, respectively, as equivalent wrenches about the closest point $\boldsymbol{p}_G$, while keeping all wrench coordinates expressed in the base frame. Accordingly, all component wrenches in $\boldsymbol{W}_n$ are represented about the same reference point and in the same coordinate frame.

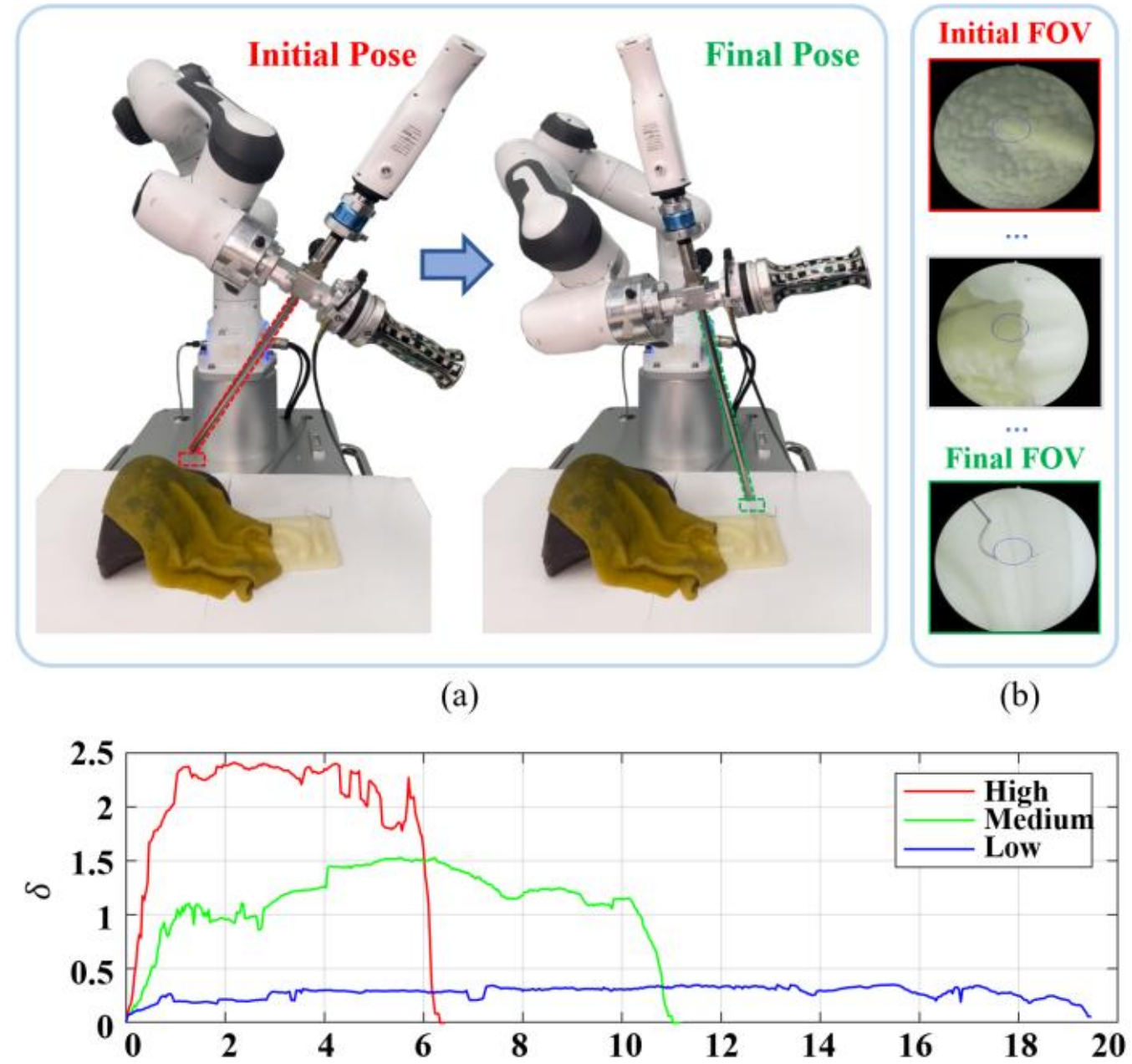

Figure 4. Compliance evaluation of the laparoscope system: (a) Initial and final poses. (b) FOV snapshots during adjustment. (c) $\delta$ profiles for three grip conditions and the corresponding completion times.

The viscous resistance wrench $\boldsymbol{\mathcal{W}}_f \in \mathbb{R}^6$ is proposed to ensure system stability. It applies viscous damping to the controlled DOFs of the laparoscope motion.

$$\boldsymbol{\mathcal{W}}_f = \begin{bmatrix}\boldsymbol{M}_f\\ \boldsymbol{F}_f\end{bmatrix} = \begin{bmatrix}-k_t(\dot{\boldsymbol{\varphi}}_G^x\boldsymbol{x}_G + \dot{\boldsymbol{\varphi}}_G^y\boldsymbol{y}_G + \dot{\boldsymbol{\varphi}}_G^z\boldsymbol{z}_G)\\ -k_f(\dot{\boldsymbol{p}}_G^x\boldsymbol{x}_G + \dot{\boldsymbol{p}}_G^y\boldsymbol{y}_G + \dot{\boldsymbol{p}}_G^z\boldsymbol{z}_G)\end{bmatrix}, \tag{40}$$

where the torque and force components of the viscous resistance wrench are denoted by $\boldsymbol{M}_f \in \mathbb{R}^3$ and $\boldsymbol{F}_f \in \mathbb{R}^3$, respectively, the positive viscous resistance parameters by $k_t$ and $k_f$, the angular and linear velocities of the laparoscope relative to the base frame expressed in frame by $\dot{\boldsymbol{\varphi}}_G$ and $\dot{\boldsymbol{p}}_G$, and the unit basis vectors of the closest point frame $\{G\}$ by $\boldsymbol{x}_G$, $\boldsymbol{y}_G$, and $\boldsymbol{z}_G$.

## IV. Phantom Experiments

To validate the effectiveness of the proposed system, phantom experiments were conducted on the experimental platform shown in Fig. 2. The model was integrated every using the Euler method to generate the reference joint positions sent to the robot.

The laparoscope assembly was mounted on the robot end-effector, and the robot controller communicated with the computer through a USB interface. The image acquisition frame rate was set to 25 fps. The laparoscope calibration method proposed by Zhang et al. [27] was used to calibrate the intrinsic parameters of a laparoscope. The YOLOv7 algorithm was employed for high-performance, low-latency surgical instrument detection.

### A. Parameter Settings

During the experiments, the parameters were set as follows. For the flexible force-sensitive sensor, the gain coefficient and

><

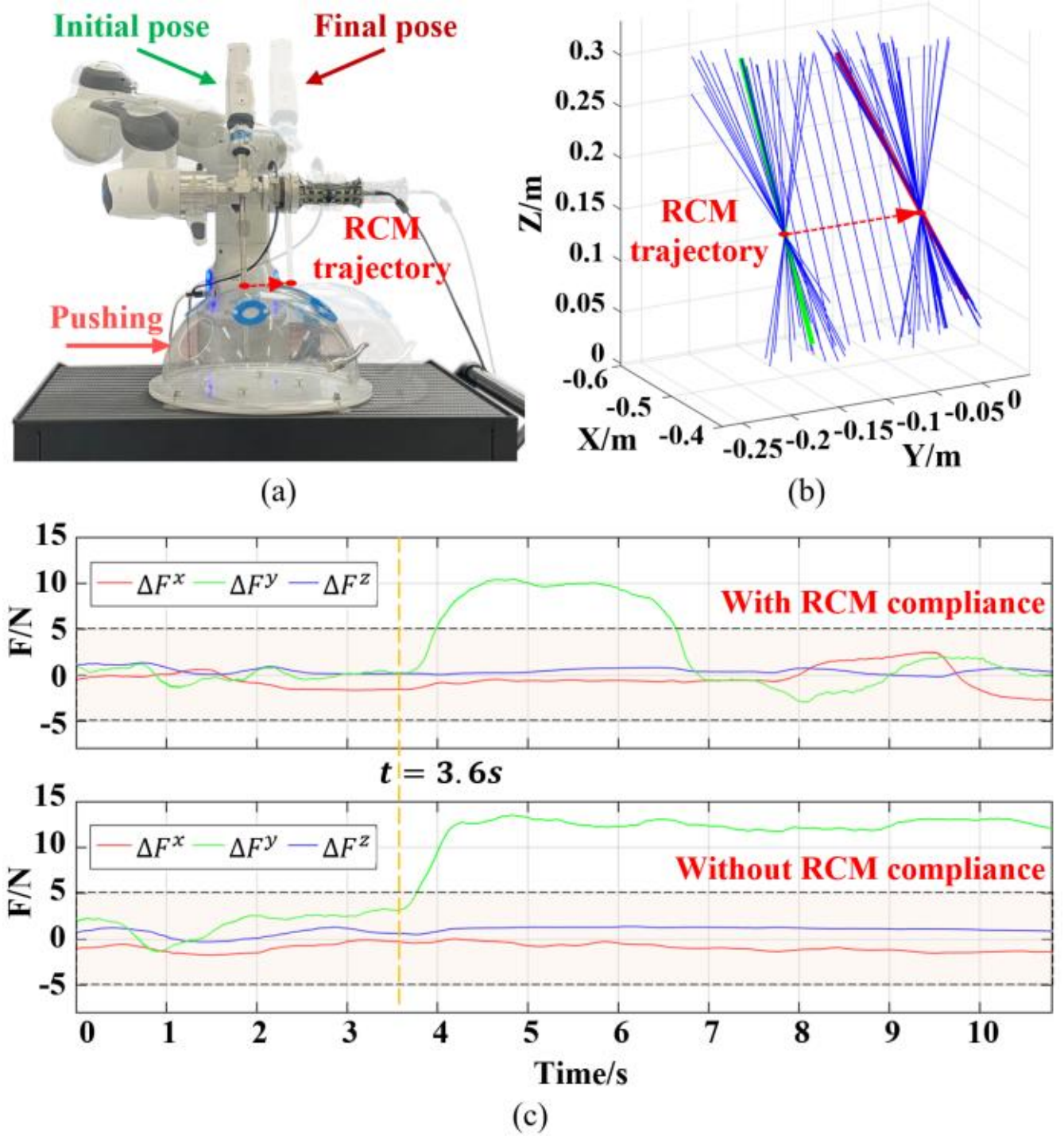

Figure 5. RCM compliance under external disturbance. (a) Initial and final configurations. (b) Laparoscope axis and RCM trajectory. (c) Incision-wall contact force $\Delta\boldsymbol{F}$ with/without RCM compliance.

the minimum gain were set to $k_H = 2000$ and $k_{\min} = 3$, respectively. For the laparoscopic camera, the intrinsic parameters were set to $u_0 = 312.22$, $v_0 = 243.39$, $f_x = 506.57$, and $f_y = 506.72$. For instrument tracking, the lower and upper distance thresholds were set to $\epsilon_1 = 50$pixels and $\epsilon_2 = 200$pixels, respectively, and the maximum tracking force was set to $F_{\max} = 10$N. In the RCM impedance law, the proportional and derivative gain matrices were set to $\boldsymbol{K}_p = 300\boldsymbol{I}_3$ N/m and $\boldsymbol{K}_d = 20\boldsymbol{I}_3$ N·s/m, respectively. In the geometric model of the laparoscope, the laparoscope length was set to $L = 0.29$m, and in the initialization of the adaptive RCM point, the scalar specifying the initial RCM location along the laparoscope was set to $\lambda_0 = 0.8$. In the dead-zone–saturation mapping of the residual force, the saturation value and the two thresholds were set to $F_{op} = 10$N, $\gamma_1 = 5.0$N, and $\gamma_2 = 15$N, respectively. In the adaptive update law of the RCM point, the adaptation coefficient, the leakage coefficient, and the update threshold were set to $\alpha = 0.003$m/(N·s), $\beta = 0.05s^{-1}$, and $\varepsilon = 5 \times 10^{-4}$m/s, respectively. Finally, in the viscous resistance wrench, the angular and translational damping coefficients were set to $k_t = 150$N·m·s and $k_f = 2$N·s/m, respectively.

### B. Compliant Laparoscope Manipulation under Different Grip Forces

This experiment evaluated the compliance of the laparoscope system under different grip forces exerted on the flexible sensor array. Compliance was quantified by the time required to achieve the same laparoscopic FOV adjustment. As shown in Fig. 4(a), the laparoscope was driven from the same initial pose to a prescribed final pose. Representative FOV images corresponding to the initial, intermediate, and final visual states are presented in Fig. 4(b), and the task was considered completed when the FOV reached the same target visual condition.

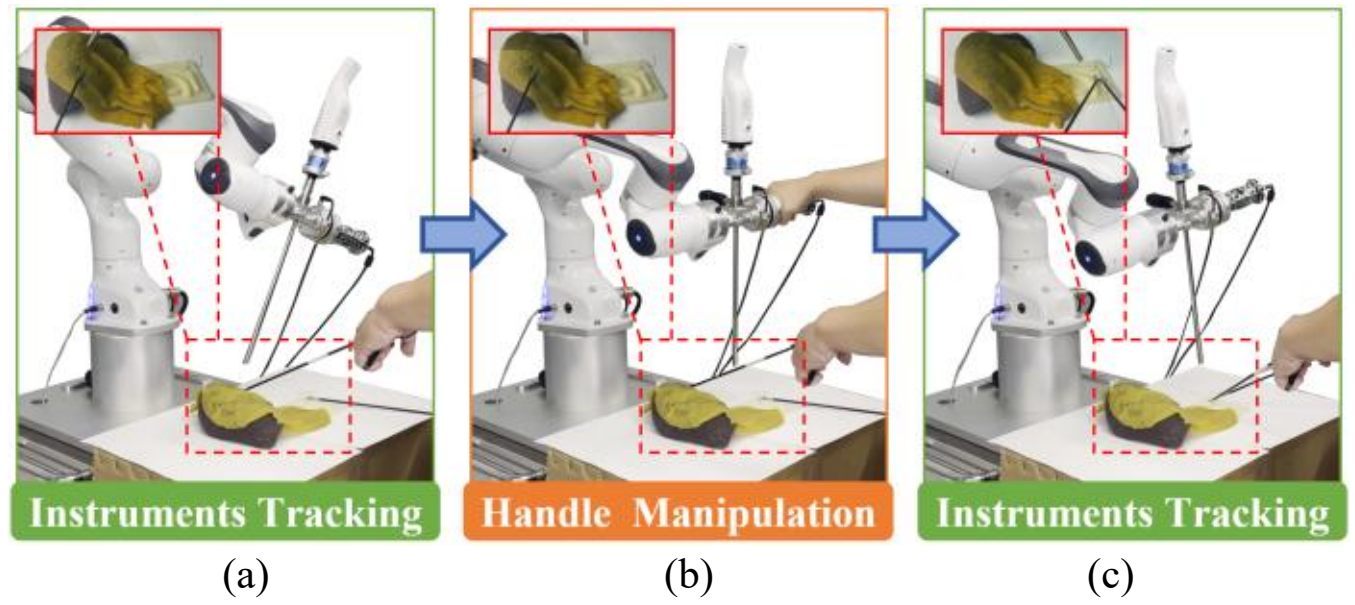

Figure 6. Experimental scenario for suture needle search: (a) Image-based surgical instrument tracking in a localized region. (b) Laparoscopic large-scale FOV adjustment controlled by handle and search for suture needle. (c) Retrieve the suture needle.

The grip intensity exerted by the operator was characterized by the sensor-derived metric $\delta$, where a larger $\delta$ indicates a stronger grip force. Fig. 4(c) depicts the time histories of $\delta$ under three grip conditions (high, medium, and low). The mean values of $\delta$ were 2.01, 1.13, and 0.28, respectively, and the corresponding completion times to reach the same FOV adjustment were 6.32 s, 11.0 s, and 19.46 s. These results demonstrate that increasing grip force (higher $\delta$) reduced the time required to accomplish the equivalent visual adjustment, thereby confirming the compliant response of the proposed laparoscope system to operator input.

### C. RCM Compliance under Incision-Site Disturbance

To evaluate the RCM compliance of the proposed system under external disturbances at the incision site, the laparoscope was operated inside a laparoscopic surgical simulator while a sustained lateral load was applied to the simulator housing to emulate incision-point displacement. The objective was to examine whether the robot could accommodate the disturbance by adjusting the RCM location and thereby alleviating the contact force between the laparoscope shaft and the incision wall. Fig. 5(a) shows the initial and final robot configurations during the disturbance. Fig. 5(b) plots the traced laparoscope axes, where the initial axis is indicated in green and the final axis in red; the corresponding evolution of the estimated RCM point illustrates that the system updated the RCM location in response to the applied disturbance, rather than enforcing a fixed pivot.

The contact force $\Delta F$exerted at the incision wall is reported in Fig. 5(c) for two conditions, namely, with and without the proposed RCM compliance, where the activation threshold was set to 5 N. The external load was introduced at $t = 3.6\,s$ (dashed vertical line). Without RCM compliance, the contact force increased abruptly and remained at an elevated level throughout the remainder of the trial, indicating persistent interaction with the incision wall. In contrast, when RCM compliance was enabled, the force exhibited a transient rise immediately after loading, followed by a gradual attenuation back toward the nominal range. These results demonstrate that the proposed RCM-constraint wrench allowed the controller to adapt the RCM location to incision-site disturbance and reduce sustained wall loading, thereby helping maintain geometric consistency and improving operational safety.

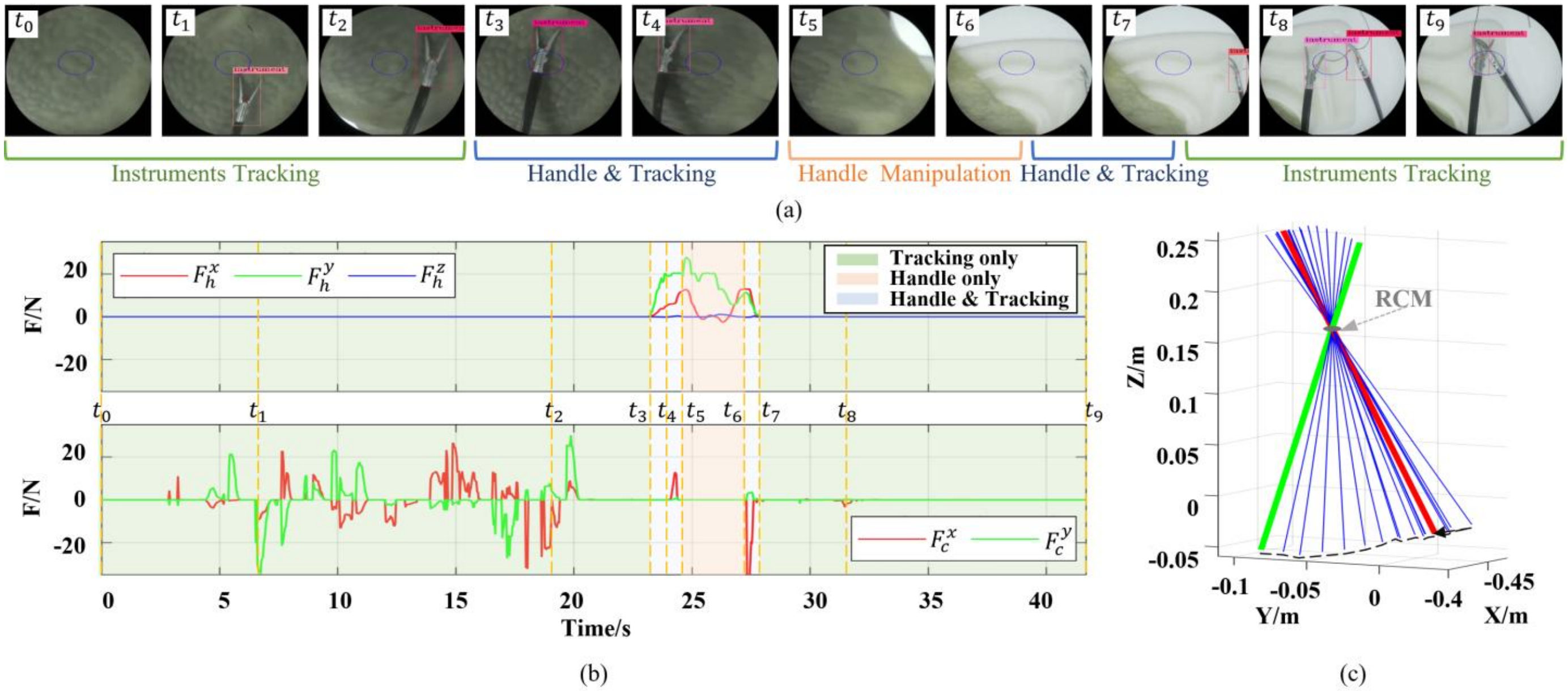

Figure 7. Experimental results for suture needle searching. (a) The experimental snapshots. (b) Applied interaction and virtual wrenches during the task. (c) Traces of the laparoscope axis: initial pose (green line), final pose (red line), fulcrum (black dot).

### *D. Cooperative Fusion of the Laparoscope-Manipulation and Instrument-Tracking Wrenches*

In minimally invasive surgery, the restricted FOV often requires substantial laparoscope reorientation to locate targets (e.g., a suture needle) that lie outside the current visual range. To evaluate the proposed control strategy in a clinically relevant scenario, a needle-search task was performed on a simulated organ phantom. The task workflow is summarized in Fig. 6: the operator first manipulates the surgical instrument within the local workspace under instrument-based visual tracking and then uses the tactile handle to command a large-scale FOV change for target searching. Once the needle is brought into view, the system returns to instrument tracking to facilitate rapid centering and subsequent retrieval.

The experimental results are presented in Fig. 7. Fig. 7(a) provides representative snapshots of the endoscopic view at key time instants $t_0$-$t_9$, with the active control mode indicated below the image sequence. Fig. 7(b) plots the measured interaction wrench applied to the handle, $\boldsymbol{\mathcal{W}}_h$, and the virtual wrench generated by the tracking controller, $\boldsymbol{\mathcal{W}}_c$. Fig. 7(c) shows the traced laparoscope axis during the task, demonstrating that the laparoscope motion remained consistent with the RCM constraint throughout the experiment.

From $t_0$ to $t_3$, the operator manipulated the instrument on the phantom, and the tracking controller generated $\boldsymbol{\mathcal{W}}_c$ to maintain the instrument in the desired image region (tracking only). Between $t_3$ and $t_5$, the operator began interacting with the tactile handle while tracking remained active (handle & tracking), producing a nonzero $\boldsymbol{\mathcal{W}}_h$ superimposed on $\boldsymbol{\mathcal{W}}_c$. At $t_5$, $\boldsymbol{\mathcal{W}}_h$ exceeded $\boldsymbol{\mathcal{W}}_c$, and the system behavior transitioned to handle-dominant control, driving the laparoscope away from the current view to perform a large FOV sweep. After $t_6$, the instrument temporarily left the image, and the operator continued handle manipulation to search the workspace until the needle appeared near the image boundary at $t_7$. The operator then released the handle, causing $\boldsymbol{\mathcal{W}}_h$ to drop, and the tracking controller re-engaged to rapidly re-establish instrument-centric viewing (from $t_7$ onward). Finally, during $t_8 - t_9$, instrument tracking was used to support precise motion, enabling successful needle retrieval.

## V. Animal Experiments

To evaluate the performance of the proposed system in a realistic surgical scenario, in vivo experiments were conducted using a porcine model. All animal procedures were reviewed and approved by the Animal Ethics Committee of Hefei University of Technology. All experiments were conducted in accordance with the relevant institutional and national guidelines for animal care and use.

The animal experimental setup is shown in Fig. 8. Surgeons from the department of thoracic surgery were invited to participate in the study. During the procedure, the surgeon manipulated the surgical instruments manually, while a 7-DOF Franka Emika robot held a wireless laparoscope. The integrated tactile handle mounted on the laparoscope was equipped with both a six-axis force/torque sensor and a flexible sensor array. The force/torque sensor was used to measure the forces and torques applied by the surgeon to the laparoscope, whereas the flexible sensor array was employed to quantify the grip tightness of the surgeon's hand on the handle. During the experiment, the proposed laparoscope-holding robotic system operated within the unified control framework, continuously providing adaptive RCM adjustment, autonomous instrument tracking, and compliant hand-guided motion. The experimental subject was a four-month-old pig weighing approximately 50 kg.

The experimental setup and representative endoscopic image sequences under two operating modes are shown in Fig. 8. On the left, the overall configuration is illustrated: the surgeon interacts with the laparoscope via a handle, the laparoscope is mounted on the robot, and the live view is displayed on a monitor during an in vivo procedure. On the right, time-stamped snapshots compare the instrument-tracking mode and the handle-manipulation mode. In instrument tracking, the laparoscope autonomously follows the surgical instruments with small, continuous view adjustments to keep the tool visible and centered. In handle manipulation, the surgeon uses the handle to reposition the laparoscope with larger and faster FOV changes for rapid target switching across anatomical regions.

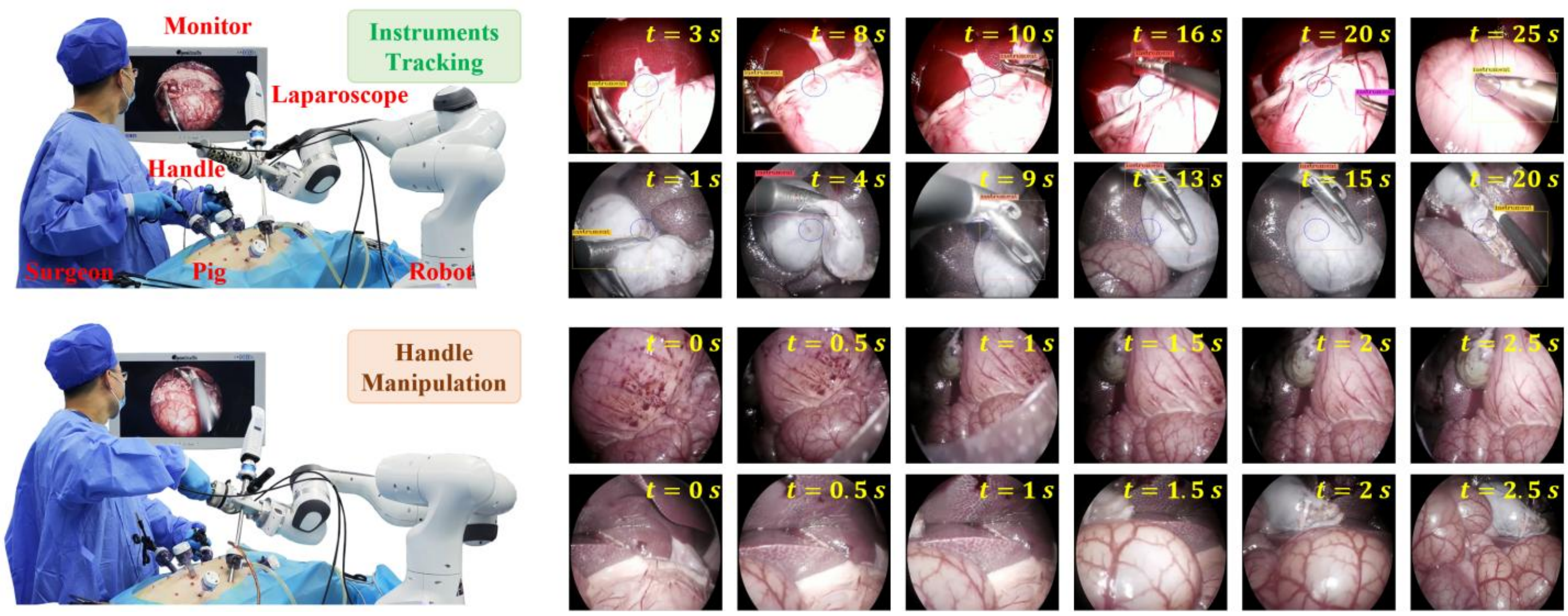

Figure 8. In vivo experimental setup and endoscopic snapshots for instrument-tracking and handle-manipulation modes.

A representative laparoscopic workflow with alternating manual handle control and autonomous visual tracking is illustrated in Fig. 9. At $t = 0\,s$ , the surgeon began manipulating the laparoscope via the handle, during which the FOV was adjusted substantially and rapidly $(t = 3\,s)$ to acquire the target region. The laparoscope was then positioned at the first gauze pad $(t = 7\,s)$, enabling the surgeon to grasp and pick it up $(t = 13\,s)$. After the target was acquired, the system switched to autonomous instrument tracking, where the FOV followed the surgical instruments with only slight adjustments $(t = 13 - 23\,s)$. Subsequently, the second gauze pad was located and retrieved under autonomous assistance $(t = 23\,s)$. The surgeon then re-engaged the handle to perform a large and rapid FOV adjustment to reposition the laparoscope toward the gallbladder $(t = 30 - 33\,s)$ . Finally, the gallbladder was exposed and removed from the abdominal cavity, completing the procedure segment$(t = 40\,s)$.

## VI. Conclusion

This paper presented a unified mechanics-based control method for laparoscope-holding robots to achieve consistent representation and cooperative fusion of multimodal intraoperative information. By mapping heterogeneous cues, including position, force/torque, and image information, into an equivalent-wrench representation in the operational space, the proposed method provides a common control-oriented framework for multimodal integration. Based on this unified representation, a task-priority scheme was developed to inject different wrenches into the task space and null space, enabling the synthesis of laparoscope control commands while coordinating multiple objectives within a single control framework. Using the intraoperative RCM position, force/torque sensing, and laparoscopic images as representative examples, we constructed an RCM-constraint wrench, a laparoscope-manipulation wrench, and an instrument-tracking wrench to support RCM-constrained motion, compliant dragging, and autonomous visual tracking, respectively. Experimental results on a surgical phantom and in vivo animal trials demonstrated that the proposed method can effectively support multi-task laparoscope operation while maintaining the RCM constraint. These results suggest that the proposed framework provides a feasible and extensible solution for multimodal laparoscope robot control in minimally invasive surgery.

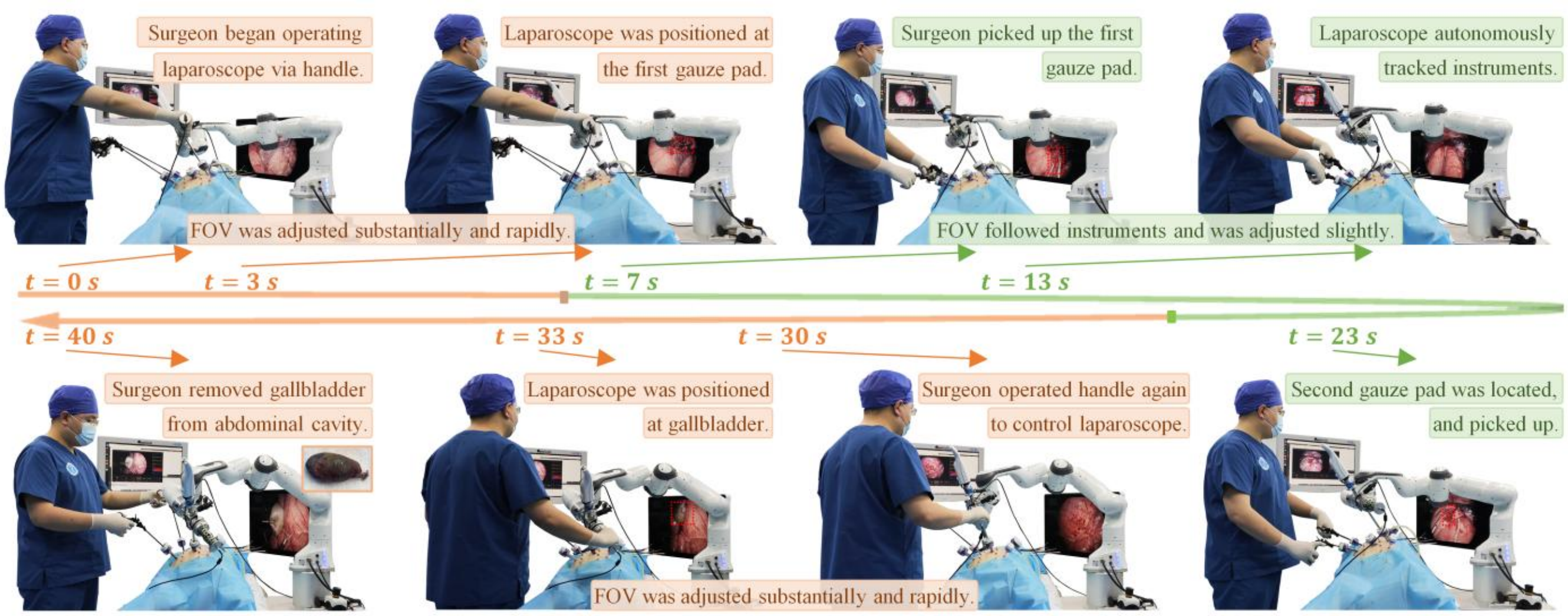

Figure 9. Representative in vivo laparoscopic workflow with alternating handle manipulation and autonomous instrument tracking.